\documentclass[10pt,twocolumn,letterpaper]{article}

\usepackage{ICCV2021/iccv}
\usepackage{times}
\usepackage{epsfig}
\usepackage{graphicx}
\usepackage{amsmath}
\usepackage{amssymb}
\usepackage{comment}

\usepackage{amsmath,amsfonts,bm}

\def\eqref#1{(\ref{#1})}

\def\1{\bm{1}}

\DeclareMathAlphabet{\mathsfit}{\encodingdefault}{\sfdefault}{m}{sl}
\SetMathAlphabet{\mathsfit}{bold}{\encodingdefault}{\sfdefault}{bx}{n}

   \usepackage{cite}

\usepackage{hyperref}
\usepackage{url}
\usepackage{hyperref}
\usepackage{url}

\usepackage{lipsum}
\usepackage{color}

\usepackage{wrapfig}
\usepackage{booktabs}
\usepackage{multirow,mathtools} 
\usepackage{caption}
\usepackage[colorinlistoftodos]{todonotes}

\DeclareMathOperator*{\minimize}{\text{minimize}}
\DeclareMathOperator*{\maximize}{\text{maximize}}

\DeclareMathOperator*{\st}{\text{subject to}}
\DeclareMathAlphabet\mathbfcal{OMS}{cmsy}{b}{n}
\newcommand{\Def}[0]{\mathrel{\mathop:}=}
\DeclarePairedDelimiterX{\inp}[2]{\langle}{\rangle}{#1, #2}

\definecolor{Sijia_color}{rgb}{0.858, 0.188, 0.478}
\newcommand{\SL}[1]{\textcolor{blue}{SL: #1}}

\usepackage{adjustbox}

 \newcommand{\add}{\mathrm{+}}
\newcommand{\remove}{\mathrm{-}}
 
\newcommand{\mycomment}[1]{}
\newcommand{\GNNViz}{\texttt{GNNViz}}
\newcommand{\TaskS}{\texttt{Task3}}
\newcommand{\TaskR}{\texttt{Task1}}
\newcommand{\TaskT}{\texttt{Task2}}

\iccvfinalcopy 

\def\iccvPaperID{10470} 

\ificcvfinal\fi

\begin{document}


\title{Preserve, Promote, or Attack? GNN Explanation via Topology Perturbation}


\author{Yi Sun\\
MIT\\
{\tt\small yis@mit.edu}
\and
Abel Valente\\
IBM Research\\
{\tt\small valente@ar.ibm.com}

\and 
Sijia Liu\\
Michigan State University\\
{\tt\small liusiji5@msu.edu}

\and
Dakuo Wang\\
IBM Research\\
{\tt\small dakuo.wang@ibm.com}
}

\maketitle
\ificcvfinal\thispagestyle{empty}\fi

\begin{abstract}
Prior works on formalizing explanations of a graph neural network (GNN)
focus on a single use case --- to \texttt{Preserve} the prediction results through identifying important edges and nodes.
In this paper, we develop a multi-purpose interpretation framework by acquiring a mask that indicates topology perturbations of the input graphs. We pack the framework into an interactive visualization system ({\GNNViz}) which can fulfill multiple purposes: \texttt{Preserve}, \texttt{Promote}, or \texttt{Attack} GNN’s predictions. 
We illustrate our approach's novelty and effectiveness with three case studies:
First, {\GNNViz} can assist non expert users to easily explore the relationship between graph topology and GNN's decision (\texttt{Preserve}), or to manipulate the prediction (\texttt{Promote} or \texttt{Attack}) for an image classification task on MS--COCO;
Second, on the Pokec social network dataset, our framework can uncover unfairness and demographic biases;
Lastly, it compares with state-of-the-art GNN explainer baseline on a synthetic dataset.
\end{abstract}


\begin{figure}[h]
\centering
\vspace*{-0.0in}
\centerline{
\begin{tabular}{c}
\hspace*{-1mm}\includegraphics[scale=0.62]{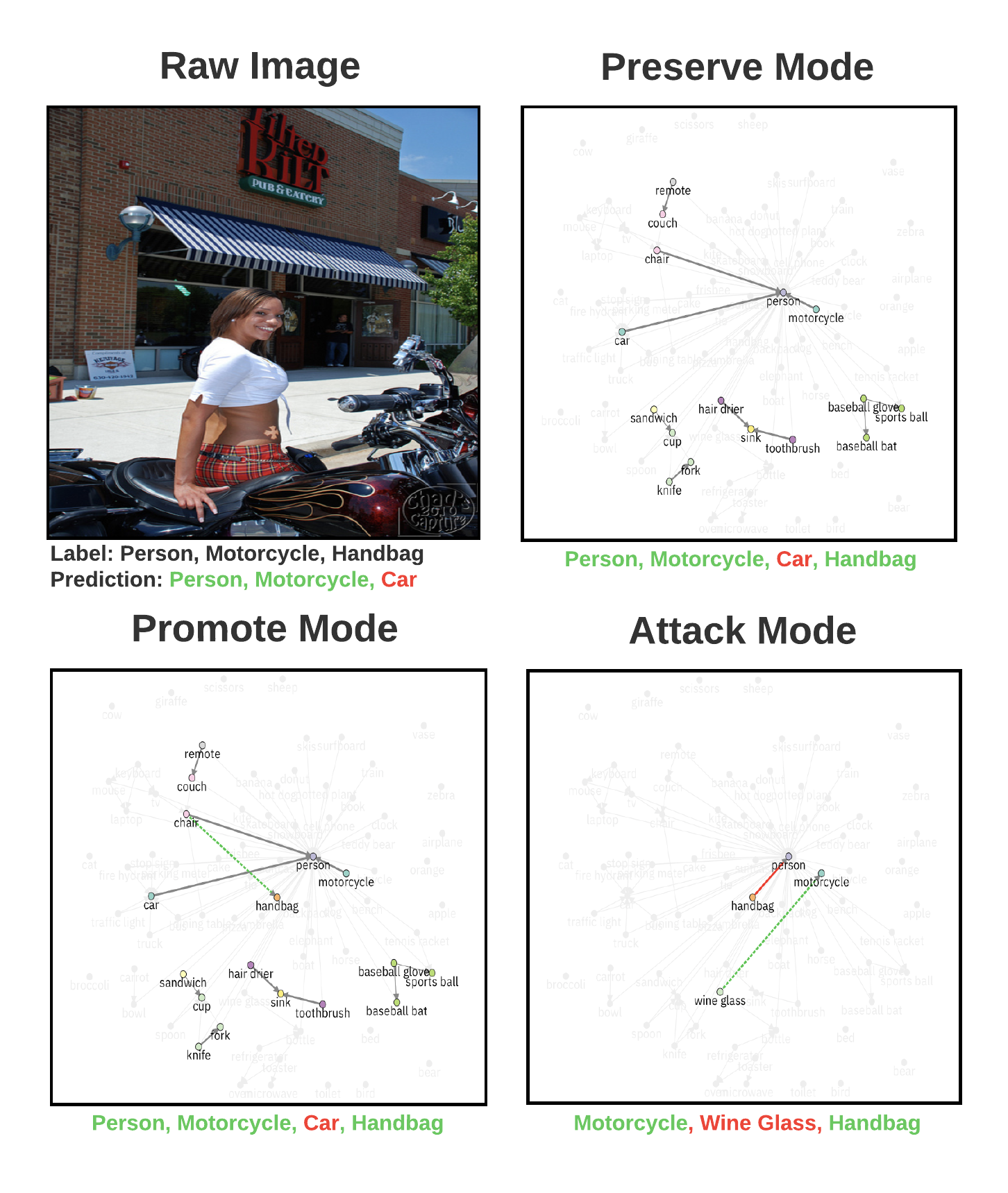}
\end{tabular}
\vspace*{-5mm}
}
\caption{{Illustration of {\GNNViz} visual analysis on MS-COCO dataset with \texttt{Preserve}, \texttt{Promote}, and \texttt{Attack}.}}
\label{fig:subgraphs-short}
\vspace*{-0.1in}
\end{figure}

\section{Introduction}

Graph neural network (GNN) is increasingly adopted in various applications for analyzing data with graph structures (e.g., social network and molecule), where the \textit{graph} can be represented as a set of entities (namely, \textit{nodes}) and
a collection of relationships (namely, \textit{edges}) between these entities ~\cite{wu2020comprehensive,zhou2018graph,kipf2016semi,velivckovic2017graph,xu2018powerful}. 
However, in real-world applications, the users of such GNN-based models also need to understand why and how a model makes certain predictions~\cite{zhang2018visual,montavon2018methods,gilpin2018explaining,xie2020explainable,bau2020understanding}.
Thus, many researchers have worked on designing various explanation mechanism for GNN-based models~\cite{baldassarre2019explainability,pope2019explainability,ying2019gnnexplainer}.

However, existing works fall short for one of the following two reasons: the graph topology as a characteristic signature of GNN has not been well studied in the GNN explanation literature; and the existing works primarily target GNN experts to analyze the model, thus they do not provide an easy-to-use tool for non-expert end users to fulfill their multi-purpose user needs. 
In this paper, we aim to answer the two research questions: \textit{Can we build an interpretation algorithm to reveal to what degree the graph topology structure affects the cause of GNN's prediction? Can we implement an interactive visualization system so that the end user can use it to interpret and alternate the GNN predictions?} 

The key to answer the first question is that, different from the existing interpretation mechanisms leveraging only the \textit{continuous} input node features \cite{baldassarre2019explainability,pope2019explainability}, 
the graph topology is typically \textit{discrete} Boolean-valued input, which poses combinatorial complexity when analyzing the sensitivity of GNN's prediction to the topology variations. 
One recent relevant work \texttt{GNNExplainer}~\cite{ying2019gnnexplainer} has to cast the GNN explanation problem with respect to the input graph as a graph learning task.
Then, identifying an explainable sub-graph is equivalent to maximizing the mutual information between the original GNN's prediction and predictions of all possible sub-graph candidates. 
However, their approach can only work for the single interpretation use case (we refer it as \texttt{Preserve} mode), but in real-world use cases, people also have the needs to change the prediction results to improve it (\texttt{Promote} mode) or to defend potential attacks (\texttt{Attach} mode). 
To accommodate all the three use cases, we tackle the GNN explanation task through a topology perturbation approach  and design a computationally-effective and multi-purpose GNN explanation framework.  

To address the second research question, we learn from the existing works on building human-in-the-loop machine learning (HITL ML) systems. 
Many interactive visualization systems have been build to interpret and debug NN models in various scenarios (e.g. face recognition \cite{masi2018deep},  automatons driving \cite{grigorescu2020survey}, and medical image analysis \cite{litjens2017survey}), so that end users can understand why the ML system made certain suggestions whether they should trust it in their decision making. 
In this work, we follow this line of work to develop the first interactive visualization system {\GNNViz} for GNN explanation, and illustrate its effectiveness in a case study with a MS--COCO classification task. Our \textbf{contributions} are three-fold:




    $\bullet$ We propose a general problem formulation and optimization framework for GNN explanation  via the  design of sparse  edge perturbations. 
    
      $\bullet$ We implement the first interactive visualization system {\GNNViz} for GNN explanation. 
      
    $\bullet$ We perform extensive quantitative and qualitative analyses with three typical GNN tasks (MS--COCO, Pokec social network, and a synthetic data) to illustrate the novelty of the proposed method over baselines.
    

\section{Related work}

DNNs are successful to achieve or even exceed
human-level performance on many  
  complex   tasks. 
   However, due to their black-box nature, it is
generally difficult to understand why DNNs make certain decisions. 
Typically, there are two approahces to explain DNN: the model-specific and model-agnostic  explanation approaches \cite{montavon2018methods,gilpin2018explaining,xie2020explainable,bau2020understanding}.
The model-specific explanation approach \cite{rudin2019stop,xie2020explainable} creates a  naturally-explainable model, such as decision trees or sparse linear or polynomial models, to mimic a DNN model as the explanation. 
Model distillation \cite{xie2020explainable} is one commonly-used strategy that learns a specific and inherently explainable machine
learning model to mimic the input-output behavior of a target black-box DNN model. 
By contrast, model-agnostic explanation methods do not creating specific interpretable models for a given DNN model. Instead, these approahces target the trained models (post hoc) \cite{bau2020understanding}. For example, 
input attribution map (IAM) \cite{adebayo2018sanity} is one well-established 
  approach, which is to 
  identify
 the aspects from the input data that drive the decisions of the DNN.
 Our research on GNN explanation follows the IAM line of work.





\vspace{-2mm}
\paragraph{DNN explanation via IAM.}
In visual recognition tasks, IAM has   been widely used to
  assign importance scores to
individual input features towards explaining a classification decision about this input.
 IAM can roughly be classified into two types:
  Jacobian saliency map \cite{simonyan2013deep,springenberg2014striving,smilkov2017smoothgrad,sundararajan2017axiomatic,zhou2016learning,selvaraju2017grad,chattopadhay2018grad}   and class-discriminative perturbation mask \cite{zeiler2014visualizing,petsiuk2018rise,fong2017interpretable,fong2019understanding}. 
  The former often calls input gradient to characterize 
the sensitivity of a model's decision to feature variations. Examples include  Vanilla Gradient \cite{simonyan2013deep}, Guided Backpropogation \cite{springenberg2014striving}, SmoothGrad \cite{smilkov2017smoothgrad},  Integrated Gradient   \cite{sundararajan2017axiomatic,zhou2016learning}, GradCAM \cite{selvaraju2017grad}, and  GradCAM++  \cite{chattopadhay2018grad}.
By contrast, class-discriminative perturbation mask directly localizes sparse  input features or image regions that maximally affect a model’s output.
Compared with Jacobian saliency map, class-discriminative perturbation mask is ease of localizing the input sparse patterns, where a decision maker focuses on.
Examples include 
Occlusion \cite{zeiler2014visualizing}, RISE \cite{petsiuk2018rise}, Mask \cite{fong2017interpretable}, and Smooth Mask \cite{fong2019understanding}.

  



\paragraph{GNN explanation.}
The GNN's input can be considered as a graph topology and its node-wise input features. 
Some recent works \cite{baldassarre2019explainability,pope2019explainability} have extended Jacobian saliency map to reveal the influence of node features on GNN's prediction.
However,  little work studied GNN explanation using the graph topology (except the recent GNNExplainer \cite{ying2019gnnexplainer}).
Since a graph topology is typically binary-valued, to use the graph topology would introduce combinatorial complexity when 
tracking the prediction change  
  by flipping individual edges on and off. Such complexity makes the construction of topological saliency map
  very challenging.  
  To walk around this challenge,
 GNNExplainer\cite{ying2019gnnexplainer} consider the GNN explanation as a task  to find the class-discriminative perturbation mask. 
 That is,  extracting a subgraph  that contributes most to a GNN's prediction by maximizing the mutual information between a GNN's prediction and predictions conditioned on possible subgraph structures.
 
 Besides the \textit{factual} GNN explanation (Preserve and Promote modes) \cite{baldassarre2019explainability,pope2019explainability,ying2019gnnexplainer}, another line of research relevant to ours is adversarial perturbation against graph topology in GNN
 \cite{zugner2019adversarial,dai2018adversarial,zugner2018adversarial,xu2019topology}. 
 It was shown  that a small amount of edge perturbations (e.g., edge addition or removal) can easily fool a well-trained GNN for an erroneous prediction. The method of generating adversarial  topology perturbation   provides a means of \textit{counterfactual} explanation (Attack mode).  It finds a similar graph tpology by changing some of the edges for which the predicted outcome of GNN changes in an adversarial way (namely, a flip in the predicted class).

In this paper, we aim to develop a unified  GNN explanation method and a visual analytics system.
Existing literature on visual analytics systems for model explanation \cite{carter2019activation,bau2017network,weidele2020autoaiviz,wang2015docuviz} also inspire our work. 
For example, Activation  Atlas  visualizes neuron activations of a convolutional neural network (CNN)  via feature inversion \cite{carter2019activation}. 
\section{Problem statement}\label{sec: problem}
In this section, 
we present the problem of 
GNN explanation by seeking topology attributions, which can
  shed light on the effect of {graph topology} (input) on 
the  GNN's  {prediction} (ouput). We also revisit the direct application of Jacobian saliency map to GNN explanation, and explore its connection to    topology attributions of our interest.

\paragraph{Preliminaries on GNN.}
Let
 $\mathcal G = (\mathcal V, \mathcal E)$ denote a graph, where $\mathcal V$ is the vertex (or node) set with cardinality $|\mathcal V| = N$, and $\mathcal E \in (\mathcal V \times \mathcal V)$ denotes the edge set. 
 The \textit{topology} of $\mathcal G$ is represented by   its adjacency  matrix 
   $\mathbf  A$, where
$A_{ij} = 1$ if $(i, j) \in \mathcal E$, and $0$ otherwise. 
Each node $i \in \mathcal V$ in $\mathcal G$ is    associated with a \textit{node feature} vector  $\mathbf x_i \in \mathbb R^{M_0}$. 
The  goal of GNN is  to learn a graph representation used for   downstream classification based on the graph topology $\mathbf A \in \mathbb R^{N \times N}$ and the node features $\mathbf X = [ \mathbf x_1, \mathbf x_2, \ldots, \mathbf x_N  ]^T \in \mathbb R^{N \times M_0}$.


Formally,  a GNN  of $L$ layers
obeys the following layer-wise propagation rule
\cite{kipf2016semi,xu2018how}
{\small \begin{align}
        \mathbf H^{(l)} = \sigma \left (  \tilde{\mathbf A} \mathbf H^{(l-1)}  \mathbf W^{(l)} 
        \right ), \quad  l \in [L],
        \label{eq: GNN_rule}
\end{align}
}%
where 
$\mathbf H^{(l)} \in \mathbb R^{N \times M_l}$ denotes the graph  representation at the $l$th layer with $\mathbf H^{(0)} = \mathbf X$,  $\sigma (\cdot )$ is the element-wise  activation function,
$\mathbf W^{(l)} \in \mathbb R^{M_{l-1} \times M_{l}}$ represents the model weights  at the $l$th layer, $[L]$ denotes the integer set $\{ 1,2,\ldots, L\}$, 
and
$\tilde {\mathbf A}$ is   a normalized adjacency matrix $\tilde{\mathbf A} = \hat{\mathbf D}^{-1/2} \hat{\mathbf A} \hat{\mathbf D}^{-1/2}$. Here  $\hat{\mathbf A} = \mathbf A + \mathbf I$, and $\hat{\mathbf D}$ is the degree matrix of $\tilde {\mathbf A}$.

A   classification head is then introduced   over the  graph representation  $\mathbf H^{(L)}$ acquired using \eqref{eq: GNN_rule} to make    node-wise or graph-wise classification decisions.
Together with labels of training data, 
a cross-entropy loss $\ell(\mathbf W; \mathbf A)$ is then minimized for seeking the optimal GNN parameters $\mathbf W = \{ \mathbf W^{(l)} \}_{l \in [L]}$.
Here for ease of presentation, we omit the dependence of $\ell$ on  node features $\mathbf X$ and training labels except its  dependence on the graph topology $\mathbf A$.
\paragraph{GNN explanation: A warm-up on   Jacobian saliency   map.}
 Jacobian saliency   map (JSM) is one of the most straightforward approaches, which assigns importance scores to
individual input features toward explaining a model's  decision. The  application of  JSM to  GNN  
has been explored   with respect to (w.r.t) node features  \cite{baldassarre2019explainability,pope2019explainability}. 
Here we revisit it from a graph topology perspective. 

Given a well-trained GNN $\mathbf W^*$ (which minimizes $\ell(\mathbf W; \mathbf A)$) and a specific input pair $(\mathbf X, \mathbf A)$, let  $p_c(\mathbf A; \mathbf W^*, \mathbf X)$ denote the prediction score of class $c$ for the learned GNN at input $\mathbf X$ and $\mathbf A$. 
The goal is 
to seek (explainable) topology attributions that can
  reveal the effect of   $\mathbf A$ on prediction $p_c(\mathbf A; \mathbf W^*, \mathbf X)$. 
  From the perspective of JSM, the topology attribution map can be cast as%
  \begin{align}\label{eq: JSM}
      \mathbf J \Def \frac{\partial p_c(\mathbf A; \mathbf X, \mathbf W^*)}{\partial \mathbf A},
  \end{align}
  where $\frac{\partial p_c}{\partial \mathbf A}$ denotes the gradient operation  w.r.t. $\mathbf A$. Besides the generation of topology attributions,
 the problem formulation and approaches proposed in our work are also applicable to generating node feature attributions of a GNN.

The rationale behind JSM   \eqref{eq: JSM} can be drawn by solving   the  following prediction maximization problem w.r.t. a topology perturbation variable $\boldsymbol \Delta$,
\begin{align}\label{eq: prob_pred_max}
    \begin{array}{ll}
   \displaystyle \maximize_{\| \boldsymbol \Delta \|_{\infty} \leq \epsilon}     &  p_c(\mathbf A + \boldsymbol \Delta ; \mathbf X, \mathbf W^*),
    \end{array}
\end{align}
where $\epsilon$ is the   radius of an $\ell_\infty$-norm based perturbation ball.
By Taylor expansion, we obtain
\begin{align}\label{eq: FO_approx}
    p_c(\mathbf A + \boldsymbol \Delta ; \mathbf X, \mathbf W^*) \approx  p_c(\mathbf A ; \mathbf X, \mathbf W^*) + \inp{\boldsymbol \Delta}{ \mathbf J},
\end{align}
where $\inp{\cdot}{\cdot}$ is the inner product operation. 
By maximizing \eqref{eq: prob_pred_max} with the first-order approximation \eqref{eq: FO_approx}, the solution is
\begin{align}\label{eq: perturbation}
    \boldsymbol \Delta = \epsilon \times \mathrm{sign}(\mathbf J),
\end{align}
where $\mathrm{sign}(\cdot)$ is the element-wise sign operation. 

The implications of JSM can then be acquired   from its connections to the topology perturbation $\boldsymbol \Delta$ via \eqref{eq: perturbation}.
{First},
the \textit{positive} entries of 
$\mathbf J$ provide a \textit{factual explanation} of edge attributions: 

$\bullet$ If $J_{ij} > 0$, namely,
$\Delta_{ij} >0$ in \eqref{eq: perturbation}, then  the promotion of  edge  attribution   from   $A_{ij}$ to $A_{ij} + \Delta_{ij}$ contributes to \textit{prediction maximization}.
 $J_{ij}$ denotes the $(i,j)$th entry of $\mathbf S$.

Second,
 $-\boldsymbol \Delta$ can   be  regarded as {adversarial perturbation} \cite{Goodfellow2015explaining,liu2018signsgd} for \textit{prediction minimization} by solving problem \eqref{eq: prob_pred_max} with the objective function $-p_c(\mathbf A + \boldsymbol \Delta ; \mathbf X, \mathbf W^*)$. Thus, the \textit{negative} entries of $\mathbf J$ provide a \textit{counterfactual explanation}:

$\bullet$ If $J_{ij} < 0$, namely, $- \Delta_{ij} > 0$ in \eqref{eq: perturbation}, then the promotion of edge attribution from $A_{ij}$ to $A_{ij} - \Delta_{ij}$ minimizes prediction accuracy, namely,
causes the vulnerability of GNN against the edge-wise adversarial attack \cite{zugner2019adversarial,dai2018adversarial,zugner2018adversarial,xu2019topology}.

Although the continuous-valued JSM \eqref{eq: JSM} provides  interesting  insights on factual and counterfactual explanation of edge attributions, it  has been shown in \cite{ying2019gnnexplainer} that JSM cannot    accurately characterize the impact of the discrete-valued graph topology $\mathbf A$ on GNN's prediction. The GNN explanation method proposed in \cite{ying2019gnnexplainer} is  most relevant  to ours. 
However, it was restricted to seeking a sub-graph of $\mathbf A$ for factual explanation only, and it called a computationally-intensive mutual information based   optimizer. 
Spurred by these limitations, we ask: \textit{Is it possible to develop a unified and computationally-efficient approach for multi-purpose GNN explanation?} We tackle this problem in next section.
\section{Methodology of multi-purpose   explanation} 

In this section we devise the topology attribution map for GNN explanation from topology perturbations perspective.

\paragraph{General idea.}
Given the binary adjacency matrix $\mathbf A$, 
(with zeros in the diagonal), 
its \textit{supplement} is given by $\bar {\mathbf A} = \mathbf 1 \mathbf 1^T - \mathbf I -   \mathbf A $, where $\mathbf I$ is an identity matrix, and 
$(\mathbf 1 \mathbf 1^T - \mathbf I)$ corresponds to the fully-connected graph.
All the possible edge perturbation candidates (addition and removal) are then given by $\mathbf C = \bar {\mathbf A} -   \mathbf A,$
where the diagonal entries of $\mathbf C $ are zeros, and its off-diagonal entries are within $\{ -1, 1\}^{N \times N}$. 
The \textit{negative} entry of $\mathbf C$ denotes the {existing} edge that can be \textit{removed} from  $\mathbf A$, and the \textit{positive} entry of $\mathbf C$ denotes the {new} edge that can be \textit{added} to the graph $\mathbf A$. Thus, $\mathbf C$ is decomposed as
\begin{align}\label{eq: C_decompose}
    \mathbf C = \mathbf C^{\remove} + \mathbf C^{\add}, ~  \mathbf C^{\remove} \leq 0, ~\mathbf C^{\add} \geq 0,
\end{align}
where $\mathbf C^{\remove}$ (or $\mathbf C^{\add}$) characterizes the non-positive (or non-negative) entries of $\mathbf C$.

Based on \eqref{eq: C_decompose},
we then introduce two Boolean \textit{edge selection {variables}}
$\mathbf S^{\remove}$ and $\mathbf S^{\add}$ associated with $\mathbf C^{\remove}$ and $\mathbf C^{\add}$, respectively. Here $\mathbf S^{\remove}$ (or $\mathbf S^{\add}$) maintains  zero entries of the same positions as $\mathbf C^{\remove}$ 
(or $ \mathbf C^{\add}$). 
And  $ S^{\remove}_{ij} = 1$  if     the edge $(i,j)$ is selected for   removing from $\mathbf A$, and $0$ otherwise. 
Similarly, $ S^{\add}_{ij} = 1$  if    the edge $(i,j)$ is selected for   adding to   $\mathbf A$, and $ 0$ otherwise.
With the aid of  $\mathbf S^{\remove}$ and $\mathbf S^{\add}$, the \textit{topology attribution map} is modeled by the {perturbed graph}
 \begin{align}\label{eq:A_indci} 
    \mathbf A^{\prime} (\mathbf S) = \mathbf A +  \mathbf C^{\remove} \circ \mathbf S^{\remove}+ \mathbf C^{\add} \circ \mathbf S^{\add}, 
\end{align}
where $\mathbf S = [ \mathbf S^{\remove}, \mathbf S^{\add} ]$ summarizes the edge manipulation operations, and $\circ$ represents the element-wise product.
Our goal is to seek the optimized $\mathbf S$ such that $    \mathbf A^{\prime} (\mathbf S)$ encodes the edges that contribute most to the  GNN's prediction.

\paragraph{Sparsity-promoting optimization.}
We cast the problem of graph explanation  as a sparsity-promoting optimization problem over the perturbation variables $\mathbf S = [ \mathbf S^{\remove}, \mathbf S^{\add} ]$,

\vspace*{-3mm}
{\small
\begin{align}\label{eq: prob_general}
      \hspace*{-5mm}  \begin{array}{ll}
   \displaystyle \maximize_{ \mathbf S }     &   \mathcal R(\mathbf A^{\prime}(\mathbf S) ; \mathbf X, \mathbf W^*) + \lambda_1 \| \mathbf S^{\remove} \|_1  - \lambda_2 \| \mathbf S^{\add} \|_1  \\
   \st & S_{ij} \in [0, 1], \forall i,j,
    \end{array}
    \hspace*{-5mm}
\end{align}
}%
\noindent where $\mathcal R$ denotes a utility function to measure the  GNN's prediction performance under   a perturbed graph $\mathbf A^{\prime}(\mathbf S)$, $\lambda_1 > 0$ (or $\lambda_2 > 0$) is a regularization parameter to strike a balance between the prediction performance and the number of removed edges (or the number of added edges), the original Boolean variable $S_{ij} \in \{ 0, 1\}$ is relaxed to its convex hull $S_{ij} \in [ 0, 1]$ for ease of optimization, and  $\ell_1$ norms are introduced in the objective function for promoting the sparsity of $\mathbf A^{\prime}(\mathbf S)$. 
In \eqref{eq: prob_general}, if $ \gamma_1$ and $\gamma_2$ increase, then the sparsity of $\mathbf A^{\prime} (\mathbf S)$ increases as the larger $\| \mathbf S^{-1}\|_1$ indicates removing more edges and the smaller $\| \mathbf S^{+}\|_1$ corresponds to adding less edges.
Since $\mathbf S$ are non-negative, $\| \mathbf S^{\remove} \|_1 = \mathbf 1^T  \mathbf S^{\remove} \mathbf 1$ and $\| \mathbf S^{\add} \|_1 = \mathbf 1^T  \mathbf S^{\add} \mathbf 1$. As a result, problem \eqref{eq: prob_general} can efficiently solved using a continuous optimization solver, e.g., projected gradient descent.


We remark that 
once the continuous optimization problem \eqref{eq: prob_general} is solved,  one can regard the resulting $\mathbf S$  as a probability matrix with elements drawn from Bernoulli distributions. Thus, 
a hard thresholding operation (e.g., compared with $0.5$) or a randomized sampling method
can be called to map a continuous solution to its discrete domain \cite{blum2003metaheuristics}. 

Moreover, in node classification task, this formulation can be adapted to provide explanations for a subset $\Theta$ containing $m$ nodes. In this case, the utility function in  problem \eqref{eq: prob_general} becomes
$
\frac{1}{m} \sum_{i=1}^m \mathcal R(\mathbf A^{\prime}(\mathbf S) ; \mathbf X_i, \mathbf W^*)
$,
where $\mathbf X_i$ indicate features of node $i$. This formulation can be particular useful for providing explanations on the group level. 

Lastly, we note that
GNNExplainer \cite{ying2019gnnexplainer} can  be regarded as a special case of  \eqref{eq: prob_general} by excluding
  the case of edge addition, namely,   imposing $\mathbf S^{\add} = \mathbf 0$. In this case, the perturbed graph
  $    \mathbf A^{\prime} (\mathbf S) $ encodes a sub-graph of the original graph $\mathbf A$ by only allowing edge removals   based on the perturbation scheme $\mathbf S^{\remove}$.

\paragraph{\texttt{Preserve}, \texttt{Promote} and \texttt{Attack} mode.}
Based  on the choice of the utility function $\mathcal R$ in \eqref{eq: prob_general}, we can use the same formulation to achieve
GNN explanation for three use cases: (1) preserving  the original predictions, (2) correcting the original predictions, and (3) attacking the original predictions. We call these three different modes \texttt{Preserve}, \texttt{Promote}, and \texttt{Attack}. 

\texttt{Preserve}
mode seeks the sparse graph pattern $\mathbf A^{\prime}(\mathbf S)$, which provides factual GNN explanation. Toward this goal, we choose 
the utility function $\mathcal R$
as the negative cross-entropy 
between the original predicted classes and the predicted classes under perturbed topology:

\vspace*{-3mm}
{\small\begin{align}
    \label{eq: preserve}
     \mathcal R(\mathbf A^{\prime}(\mathbf S) ; \mathbf X, \mathbf W^*) = &
       \sum_{c \in \Omega}y_c\log p_c(\mathbf A^{\prime}(\mathbf S) ; \mathbf X, \mathbf W^*),
\end{align}}%
where $\Omega$ is the set of original prediction labels. 

\texttt{Promote} mode aims to fix GNN's wrong predictions. Mathematically, the goal is to seek a sparsity-promoting topology $\mathbf A^{\prime} (\mathbf S)$ so under this perturbed topology, the model predicts the desired set of prediction labels with higher probability compared to other alternative labels. This goal can be  modeled with  C\&W-type objective function  \cite{carlini2017towards}:

\vspace*{-3mm}
{\small
\begin{align}
    \label{eq:  CW_loss_fact}
     \mathcal R(\mathbf A^{\prime}(\mathbf S) ; \mathbf X, \mathbf W^*) = &
       \min  \left  \{  \min_{c \in \Omega} p_c(\mathbf A^{\prime}(\mathbf S) ; \mathbf X, \mathbf W^*)  \right. \nonumber \\
       & \left.- \max_{t  \notin \Omega}  p_t(\mathbf A^{\prime}(\mathbf S) ; \mathbf X, \mathbf W^*),   \kappa  \right \},
\end{align}}%
\noindent where $\Omega$ is the set of desired prediction labels, and $\kappa \geq 0$ is a given confidence level. The rational behind \eqref{eq:  CW_loss_fact}  is that  the maximization of $\mathcal R$ would force the  margin of correct prediction $ \min_{c \in \Omega} p_c(\mathbf A^{\prime}(\mathbf S) ; \mathbf X, \mathbf W^*) - \max_{t  \notin \Omega}  p_t(\mathbf A^{\prime}(\mathbf S) ; \mathbf X, \mathbf W^*) $    to  reach the confidence level $\tau$. This also implies  
$  p_c(\mathbf A^{\prime}(\mathbf S) ; \mathbf X, \mathbf W^*) \geq   p_t(\mathbf A^{\prime}(\mathbf S) ; \mathbf X, \mathbf W^*) $ for any $c \in \Omega$ and $t \notin \Omega$, namely, the obtained $\mathbf A^{\prime} (\mathbf S)$ promotes the set of desired labels.
Note that this formulation can also be used for the purpose of preserving predictions, though we observe \eqref{eq: preserve} is more straightforward and performs better in practice. 

{\texttt{Attack} mode} 
 aims for counterfactual explanation: We seek the most sensitive edges to adversarial attacks, thus altering them would successfully fool a GNN. We thus modify the objective function \eqref{eq:  CW_loss_fact} to
 
 \vspace*{-3mm}
 {\small 
\begin{align}\label{eq:  CW_loss_nodei_atk}
 \mathcal R(\mathbf A^{\prime}(\mathbf S) ; \mathbf X, \mathbf W^*) 
= &
\min \left   \{    \max_{t \notin \Omega}  p_t(\mathbf A^{\prime}(\mathbf S) ; \mathbf X, \mathbf W^*) \right. \nonumber \\
& \left. - \max_{c \in \Omega} p_c(\mathbf A^{\prime}(\mathbf S) ; \mathbf X, \mathbf W^*),   \kappa  \right  \}.
\end{align}
}%
In contrast to \eqref{eq:  CW_loss_fact}, the maximization of \eqref{eq:  CW_loss_nodei_atk} forces
$ \max_{t \notin \Omega}  p_t(\mathbf A^{\prime}(\mathbf S) ; \mathbf X, \mathbf W^*) \geq   ( p_c(\mathbf A^{\prime}(\mathbf S) ; \mathbf X, \mathbf W^*) + \kappa )$ for any $c \in \Omega$. This implies that under the perturbed graph  $\mathbf A^{\prime}(\mathbf S)$, there exists an incorrect prediction $t \notin \Omega$ with $\kappa$ higher confidence over all the true labels  $ \{ c \in \Omega \}$.

\section{Experiments and quantitative analysis
}

In this section, we set up experiments to
quantitatively analyze the empirical performance of the  topology attribution map $\mathbf A^{\prime}(\mathbf S)$ obtained by solving problem \eqref{eq: prob_general}.
We show how $\mathbf A^{\prime}(\mathbf S)$ characterizes an explanatory topology (mask on the original graph) can encode essential  edges to \texttt{preserve, promote,} or adversarially altering the prediction (\texttt{attack}) a well-trained GNN.

\subsection{Experiment tasks setup}

\paragraph{{\TaskR}: GCN-based image classification   .}
We use the image classification task on MS--COCO dataset \cite{lin2014microsoft} as an illustration, and a well-cited GCN-based multi-label classification model \cite{Chen2019MultiLabelIR} as the target model for explanation, but our explanation framework and {\GNNViz} system are applicable to other GNN models and datasets as well. Unlike previous approaches that treat each object classifier as independent, \cite{Chen2019MultiLabelIR} proposes to use a graph convolutional network (GCN)  to capture the label correlations (Appendix Fig.\ref{fig:gcn-graph}). The ground-truth adjacency matrix is generated by computing the co-occurence probability of labels in the training set. We evaluate the explanation results of the three modes on a subset of validation dataset with 5500 images, used in\cite{Chen2019MultiLabelIR}. 

\paragraph{{\TaskT}: Bias detection in a social network dataset.} Biases in social network stems from the fact that people tend to connect with those sharing similar sensitive characteristics, and the use of GNN could further reinforce such biases. 

Pokec \cite{Takac2012DATAAI} is a popular online social network dataset with millions of user profile data in Slovakia. The profile data contains gender, age, hobbies, interest, professions, which is a good candidate to examine biases existed in GNN. We randomly extract a subset of 67,796 nodes from the Pokec dataset based on a user's region information. The target model's prediction task is to predict the professions of the users \cite{Dai2020SayNT}, we treat \textbf{region} (e.g., low-income, high-income) as a sensitive attribute.  

\paragraph{{\TaskS}: Benchmark on synthetic dataset.}
To benchmark the quality of explanations, we aslo use the node classification synthetic datasets defined in \cite{ying2019gnnexplainer}. Each dataset is constructed by attaching meaningful motifs (house, cycle, grid) to a synthetic base graph, and the motifs are treated as ground-truth topology attribution map. Note this dataset can only benchmark the \texttt{preserve} mode explanation, whereas the \texttt{promote} and \texttt{attack} modes are not investigated in prior studies. 


\paragraph{Implementation details.}
During learning of the explanation masks,
we adopt Adam as the optimizer running over  $200 $  steps  with learning rate $0.1$. 
For {\TaskS}, we choose the regularization parameters $\lambda_1 = 0.005$ and $\lambda_2 = 0$,  and the confidence tolerance $\kappa = 0.4$. For {\TaskR} and {\TaskT}, we use $\kappa = 0.2$ for all three modes, $\lambda_1=0.001$ and $\lambda_2=0.0005$ for \texttt{promote} and \texttt{attack} modes.

In order to evaluate the explanation mask, we only keep the top edges in the attribution mask. In \texttt{preserve} mode, we keep the top 10 edges in the final explanation mask. In \texttt{promote} mode, we seek to make the prediction correct by allowing a maximum of 5 edge additions. For \texttt{attack} mode, we allow a maximum of 5 edge removals and 5 edge additions to make top-1 label wrong.

\subsection{{\TaskR}: GCN-based image classification}

\paragraph{Evaluation metrics.} Although ground-truth explanations are usually unavailable, we can still quantitatively measure the effectiveness of the explanation methods by defining following evaluation metrics:

$\bullet$ \textbf{Accuracy (ACC) }: For \texttt{Preserve} mode, we measure accuracy as the consistency between prediction on the whole graph and the explanation subgraph. For \texttt{promote} and \texttt{attack} mode as the mean average precision (mAP) of the top 3 predicted labels. 

$\bullet$ \textbf{Explanation Quality (EQ)}: For the multi-label classification problem, explanation quality captures whether the explanation mask could identify edges between relevant labels. Further detailed formulation are in Appendix.

\paragraph{Experiment results.}
In Table-\ref{tab:real_results}, we present the effectiveness of our algorithm on {\TaskR}. As we can see, all of the three modes are successful in achieving their respective goals.\texttt{Preserve} mode maintains $95.67\%$ of the original prediction. The original model prediction reaches mAP of $83.0\%$, and the \texttt{promote} and \texttt{attack} mode is able to increase and decrease mAP, respectively. 

\begin{figure}[h!]
\vspace*{-0.1in}
    \centering
    \includegraphics[scale=0.25]{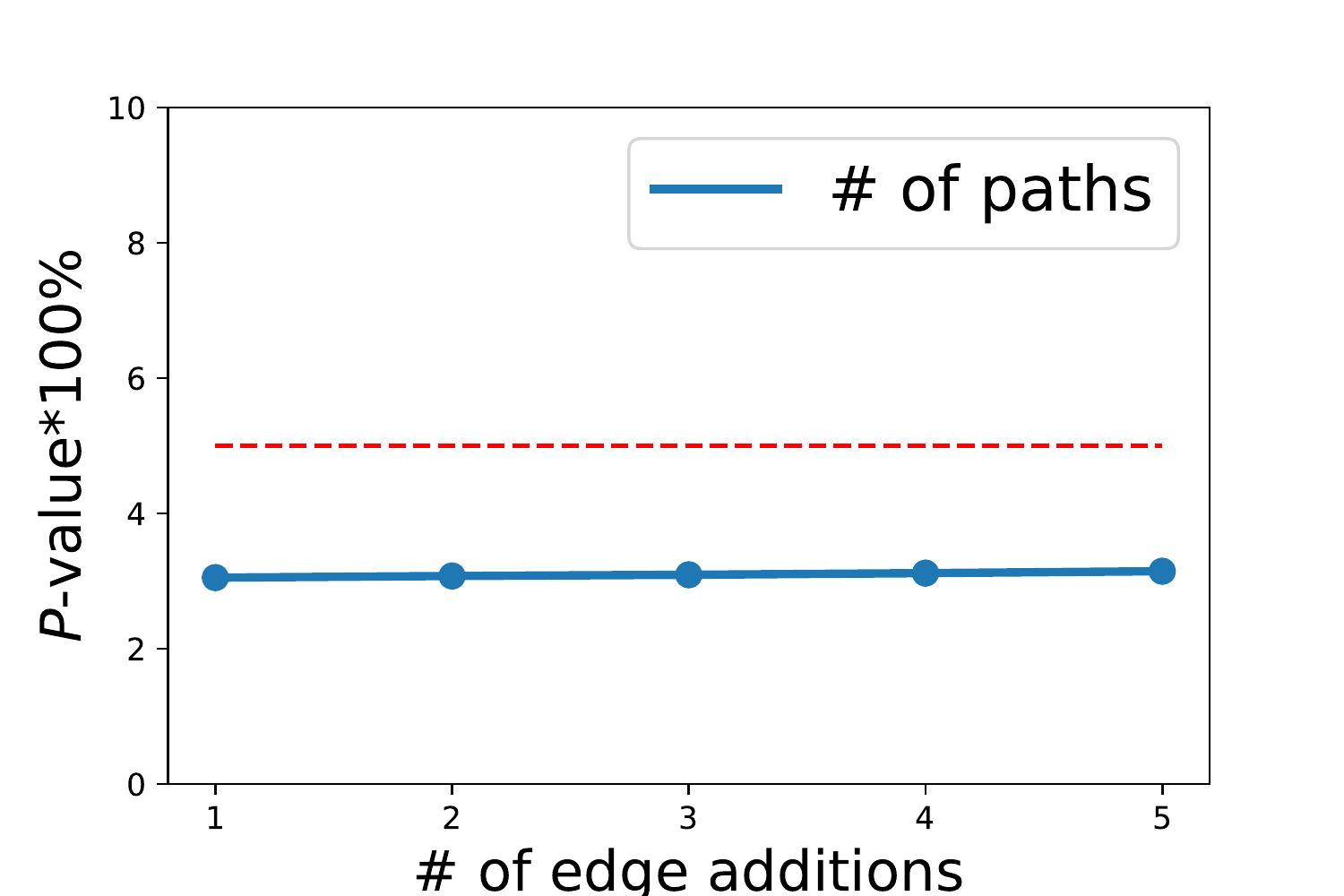}
    \includegraphics[scale=0.25]{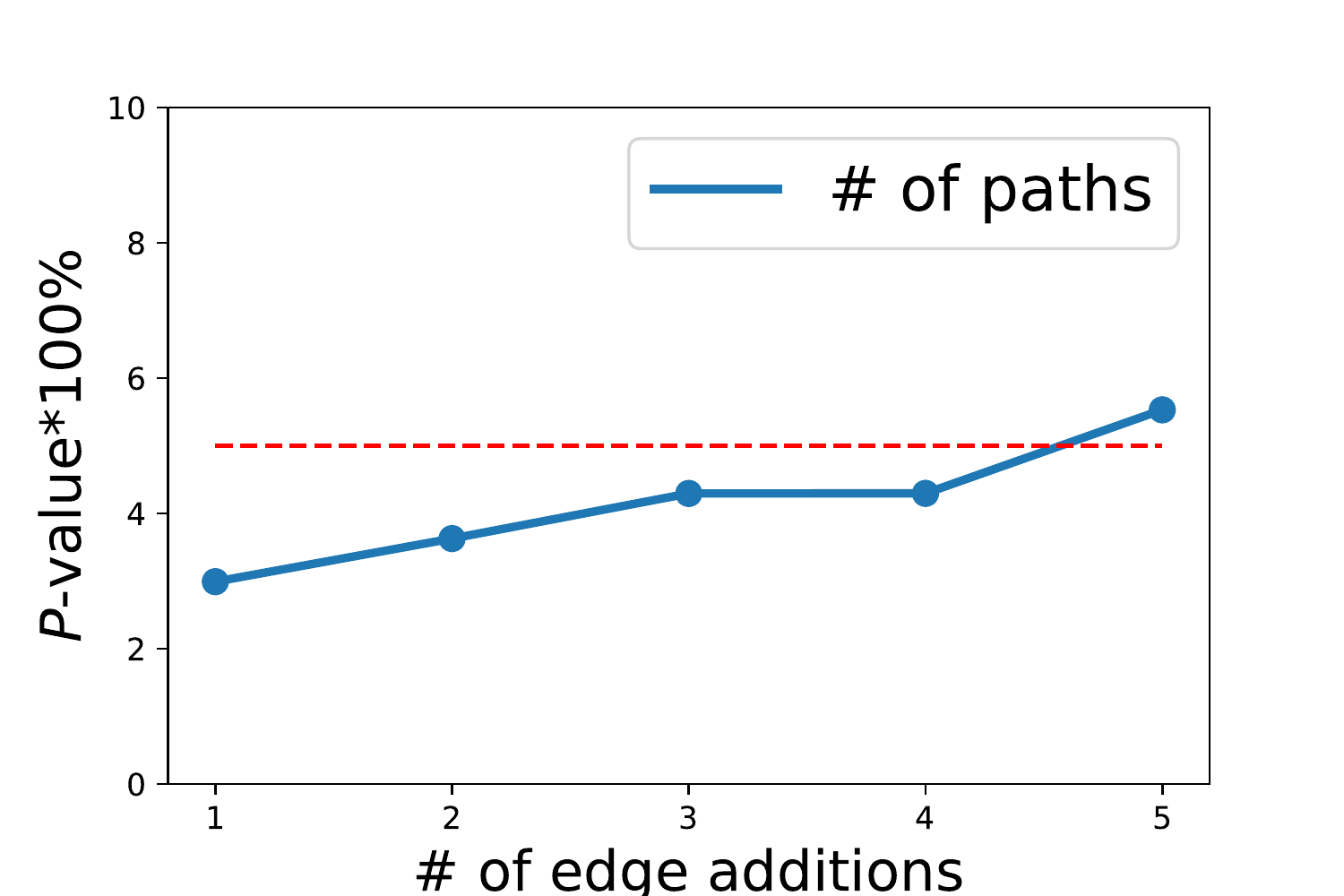}
    \caption{
    The x-axis is the number of edge additions, and y-axis is the $P$-value $\times 100\%$. The dashed line indicates statistically significance line ($P$-value=$0.05$). the Left: 0-hop neighborhood (immediate neighbors) of true labels. Right: 1-hop neighborhood of true labels. The result demonstrates that the edge addition proposed by our method statistical significantly increases connectivity between true labels.
    }
    \vspace*{-0.1in}
    \label{fig:p_value}
\end{figure}

In terms of explanation quality, the sparse explanation graph in \texttt{preserve} is able to identify $62.8\%$ of the edges connecting predicted labels. Figure-\ref{fig:p_value} demonstrates that edge additions proposed by \texttt{promote} mode statistical significantly increase connectivity between true labels.

\begin{table}
\begin{center}
\resizebox{0.45\textwidth}{!}{
\begin{tabular}{ c c c c }
\toprule
Mode & \texttt{Preserve} & \texttt{Promote} & \texttt{Attack}\\
\midrule
Accuracy & $95.67\%$  & $90.36\%$ & $25.18\%$ \\
 \bottomrule
\end{tabular}
}
\end{center}
\vspace*{-0.1in}
\caption{Quantitative results for {\TaskR}. For \texttt{preserve}, accuracy measures the mAP w.r.t the original predicted labels. For \texttt{promote} mode and \texttt{attack}, accuracy measures the mAP w.r.t the true labels.}
\label{tab:real_results}
\end{table}

\subsection{{\TaskT}: Bias detection in social network dataset}
 GNN's prediction may have fairness issues in a social network dataset. Can interventions (adding artificial edges) correct GNN's prediction biases? We show the using \texttt{promote} explanation mode can be used to address this question.

\paragraph{Evaluation metrics.}
We use the absolute difference of Equalized Odds ($\Delta EO$) \cite{Hardt2016EqualityOO} to measure the fairness loss of GCN's prediction. Let $s \in \{0,1\}$ denotes the binary sensitive label, $y$ and $\hat{y}$ indicates the true label and the predicted label respectively. Fairness loss can be measured as:
{\small
$$\Delta EO = |\mathbb{P}(\hat{y}=1|y=1,s=0)-\mathbb{P}(\hat{y}=1|y=1,s=1)|$$
}
After training for 200 epoches, the GCN model is able to achieve 71.3\% test accuracy, with $\Delta EO = 0.103$, indicating that there is a discrepancy when predicting users come from the majority group and the minority group.

\paragraph{Experiment results.}
\begin{figure}
    \centering
    \includegraphics[scale=0.45]{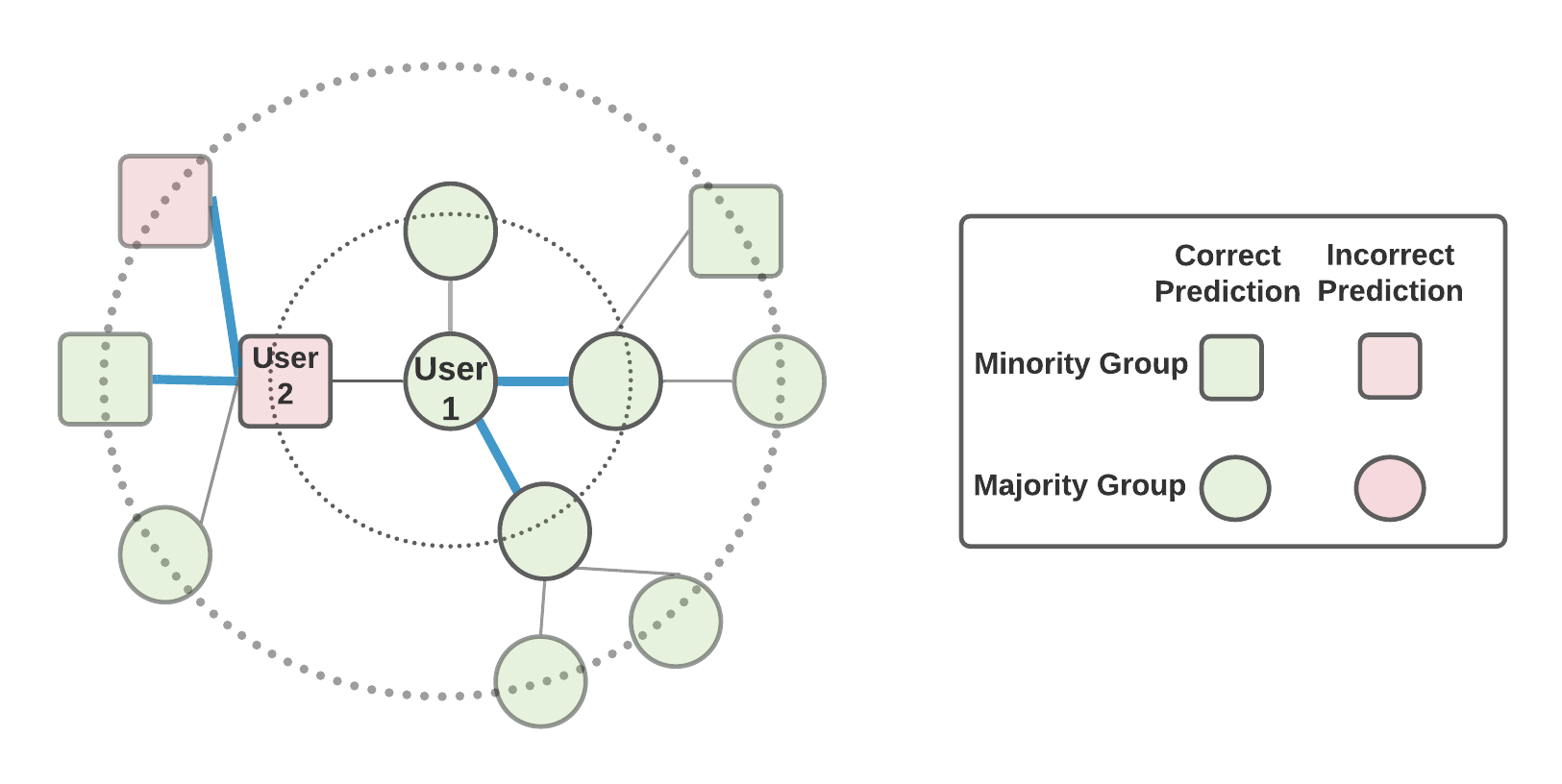}
    \caption{An example graph identifying the sources of biases in Pokec dataset. For both user 1 (majority group) and user 2 (minority group), their intra-group connections are highlighted in the explanation mask.}
    \label{fig:fairness_graph}
\end{figure}
We first illustrate how our explanation framework helps to identify the source of biases existed in GNN's prediction. In Figure-\ref{fig:fairness_graph}, a green node indicates GNN predicts the node's label (\texttt{profession}) correctly, and a red node indicates a wrong prediction. The shape of a node indicates their group (square represents the minority class based on the \texttt{region} sensitive attribute). As shown in the graph, there are more intra-group edges than inter-group edges. In addition, intra-group edges are also more important in GNN's prediction and more likely to be highlighted in the explanation mask. This further exaggerates GNN's performance discrepancy on majority group and minority group.

In Table-\ref{tab:pokec_results}, we also show how accuracy and equalized odds changes as we add more edges to the adjacency matrix. As expected, as we add more inter-group edges to network, the discrepancy of equalized odds decrease with a slight decrease in accuracy.  
\begin{table}
\begin{center}
\resizebox{0.45\textwidth}{!}{
\begin{tabular}{ c c c c c c}
\toprule
\# edges & n=50 & n=100 & n=500 & n=1000 & n=5000\\
\midrule
ACC & 0.713  & 0.712 & 0.712 & 0.703& 0.676 \\
 \midrule
$\Delta EO$   &  0.103 & 0.101  & 0.097 & 0.093& 0.052\\
 \bottomrule
\end{tabular}
}
\end{center}
\vspace*{-0.1in}
\caption{Experiment results on {\TaskS}. ACC and $\Delta EO$ as the number of added edges increases in adjacency matrix.}
\label{tab:pokec_results}
\end{table}


\subsection{{\TaskS}: Benchmark on synthetic dataset}
\begin{figure}[h]
    \includegraphics[scale=0.25]{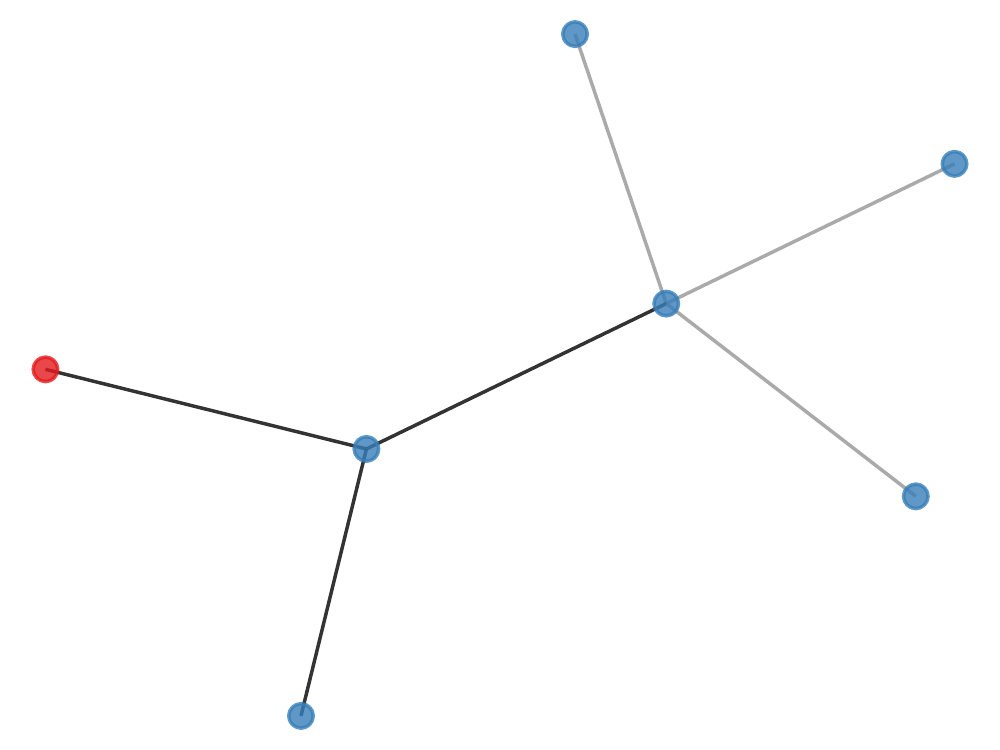}
     \includegraphics[scale=0.25]{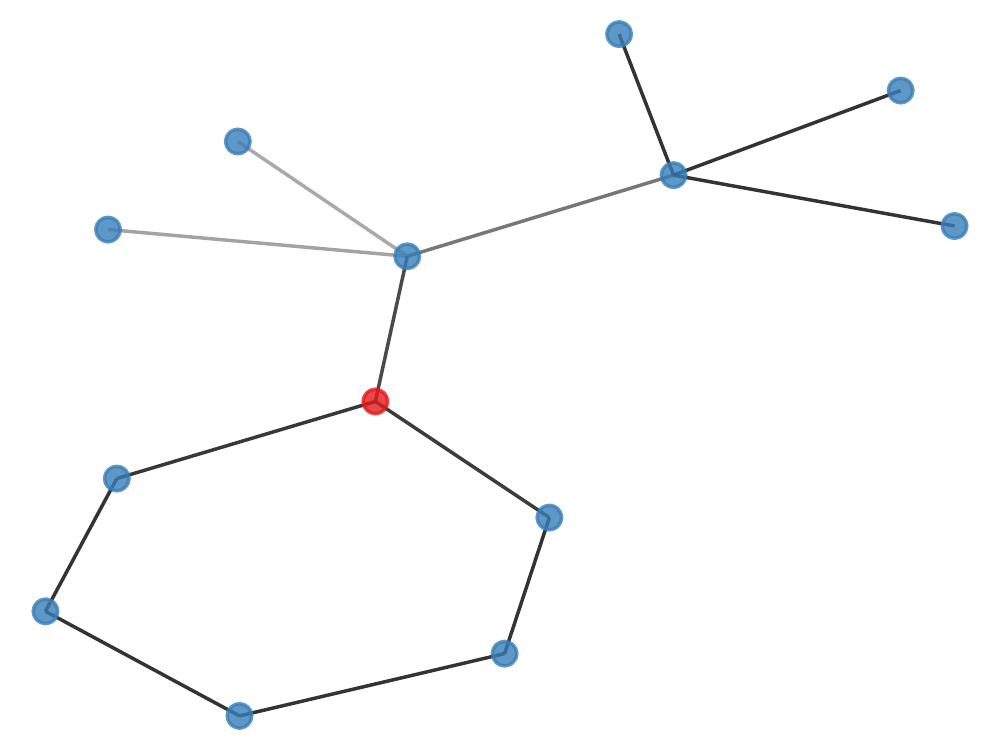}
     \includegraphics[scale=0.25]{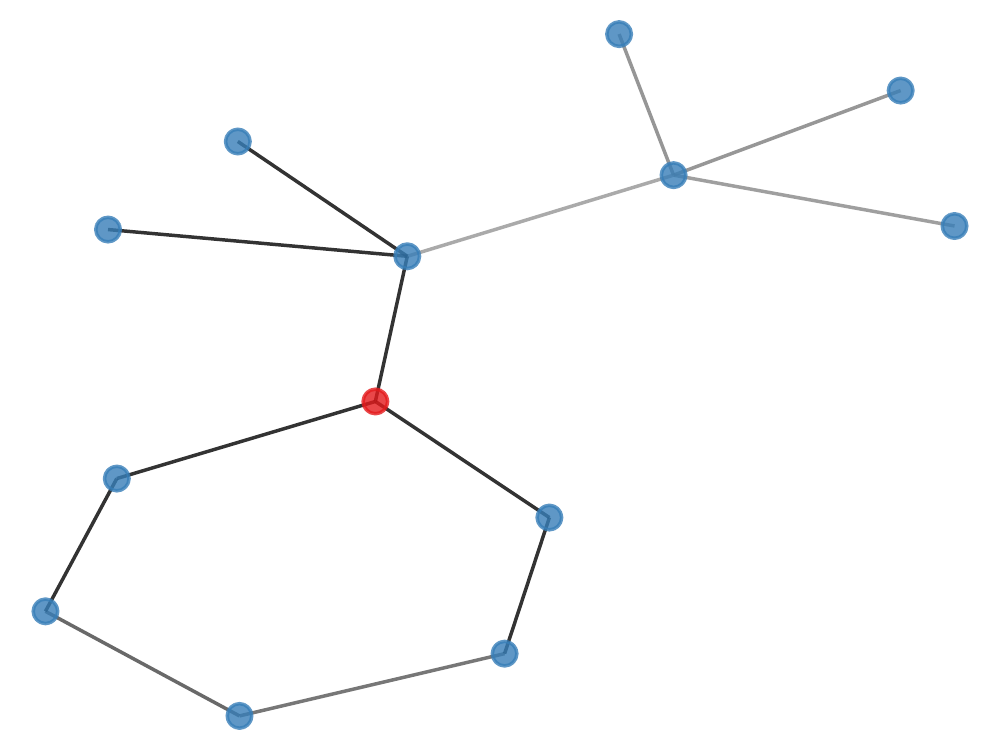}
    \caption{Example of identified cycle-shaped motifs for \texttt{Tree-Cycles}.Left: Grad. Middle: GNNExplainer. Right: Ours. }
    \label{fig:syn4}
    \end{figure}

Lastly, the evaluation experiment in {\TaskS} is a classification task, where the predicted scores are the importance weights in the explanation mask of the ground-truth graph \cite{ying2019gnnexplainer}. The explanation accuracy is measured by the AUC score of the classification task. 
We compare the explanation accuracy with two baselines: gradient-based saliency map (Grad) and GNNExplainer \cite{ying2019gnnexplainer}. Note that both baselines only consider factual explanation, thus we are only able to compare accuracy in \texttt{preserve} mode. As shown in Figure-\ref{fig:syn4}, and Table-\ref{tab:syn_results}, Our method is able to correctly identify the motifs and achieve similar explanation accuracy as GNNExplainer and significantly outperforms Grad. Although its accuracy is slightly worse in comparison to GNNExplainer, but our algorithm is not optimized solely for this single metric and \texttt{preserve} mode, thus our approach can provide a satisfying performance on this metric while being more flexible in supporting additional use cases and insights.  
    
\begin{table}[h!]
\begin{center}
\setlength{\tabcolsep}{1.5pt}
\resizebox{0.45\textwidth}{!}{
\begin{tabular}{ cccc }
\toprule
Method & \texttt{BA-Shapes} & \texttt{Tree-Cycles} & \texttt{Tree-Grid}\\
\midrule
 Grad & $0.882$ & 0.905 &  0.667 \\
 \midrule
 GNNExplainer & $0.925 $ & 0.948 & 0.875\\
 \midrule
 Ours & $0.902$ & 0.921 & 0.836  \\
 \bottomrule
\end{tabular}
}
\end{center}
\vspace*{-0.1in}
\caption{Explanation accuracy on synthetic dataset with different motifs.}
 \label{tab:syn_results}
\end{table}

 \section{Interactive visualization system {{\GNNViz}} and qualitative visual analytics}

\begin{figure*}[htb]
    \centering
    \includegraphics[scale=0.6]{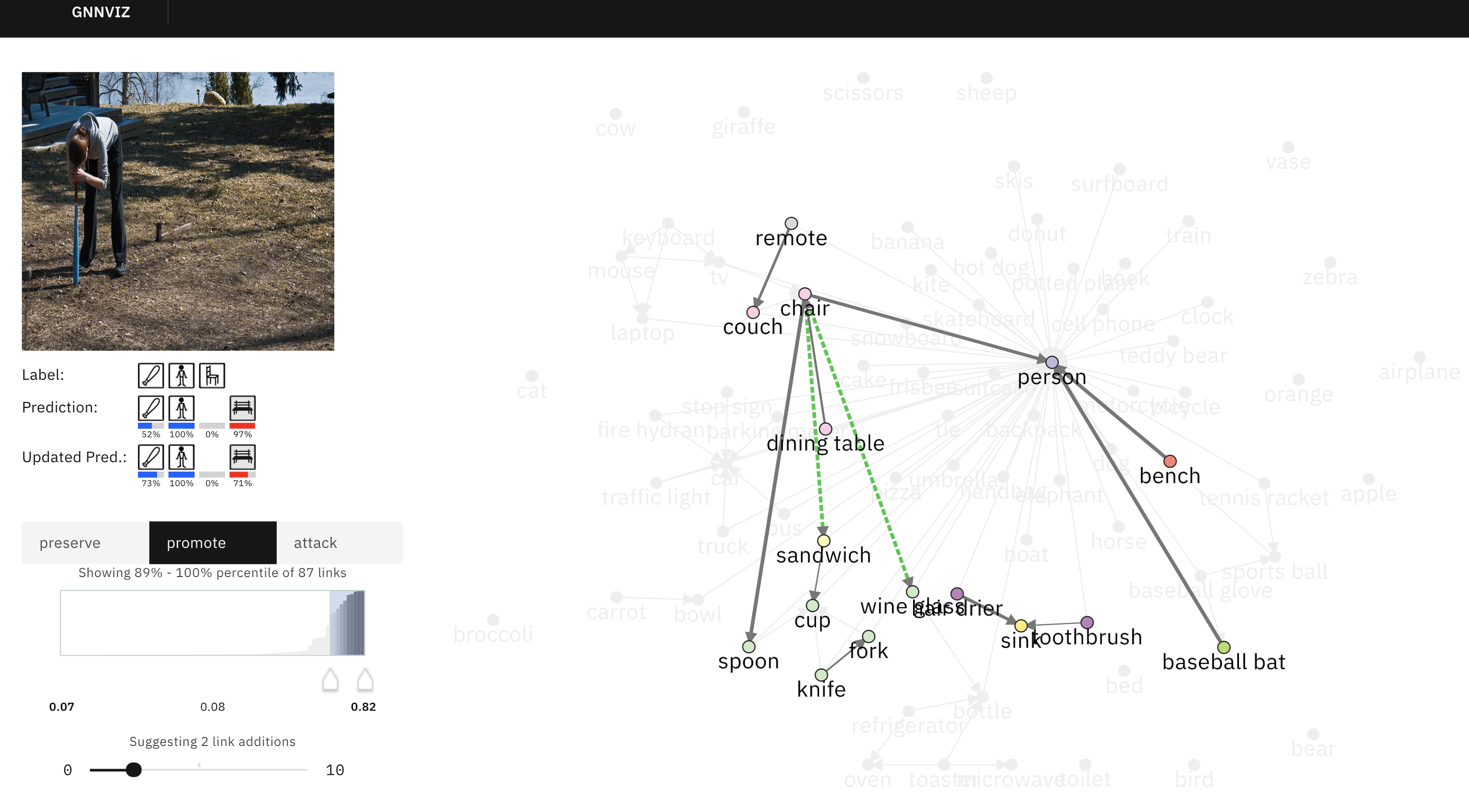}
    \caption{ {\GNNViz} System on MS--COCO data analysis. To the top-left, the system shows the image input, the groundtruth label, prediction label, and the updated labels when user changes the graph topology. To the bottom-left, it allows users to alternate among the three models (currently on Promote mode). To the right, it shows the identified subgraph (solid black lines) on top of the global graph (translucent), and the suggested two links to add in the Promote mode (in green dash lines).}
    \label{fig:ui-frontpage}
\end{figure*}

We build {\GNNViz} system to enable non-expert end users to easily leverage on our explanation algorithm to fulfill their multiple user purposes.  As shown in Fig.\ref{fig:ui-frontpage}, the system has functionalities that allow a user to quickly navigate through the data points of a dataset, examine the data's groundtruth labels, and interpret the GNN predictions in three modes. A more detailed description is in Appendix. 

In this section, we use only {\TaskR}: GCN-based image classification on MS--COCO~\cite{Chen2019MultiLabelIR} as an illustration of how the end users can perform qualitative analytics in three modes with {\GNNViz} system, but our system can support other GNN models and datasets as well.
The majority space of UI is dedicated to the global graph topology. The global graph topology map in this case is calculated from the label co-occurence probability from the entire training dataset. Thus, each node in the graph represents a label, and each directed edge represents an co-occurence relationship. The edges are binary, with a threshold of conditional probability at 0.4, as suggested by the original paper~\cite{Chen2019MultiLabelIR}. 

The default user mode is \texttt{preserve} mode. The \texttt{preserve} mode falls into the factual explanation use case -- the explanation algorithm identifies a less complex subgraph for prediction, but still try to maintain the best prediction performance as using the original global graph topology. 
The second user mode is \texttt{promote} mode, and it is illustrated in Fig.\ref{fig:ui-frontpage}. The \texttt{promote} mode is also for the factual explanation user scenario. Our {\GNNViz} algorithm can suggest new edges to add into the graph topology and promote the correct prediction results. 
The third user mode is \texttt{attack} mode. In contrary to the previous two user modes, this mode is to visualize the counterfactual explanation user scenarios. {\GNNViz} can recommend the minimal edge changes (both edge addition and deletion) to attack the GCN prediction algorithm -- manipulate the updated prediction results to move away from the correct predictions.
It uses green dash lines to indicate the edge addition, and uses red solid lines to indicate edge removal, as shown in Fig.~\ref{fig:subgraphs-short}.

The layout of this graph (Fig.~\ref{fig:system-ui} \textbf{C}) uses a force-directed graph layout~\cite{fruchterman1991graph} and the implementation uses a D3.js javascript library~\cite{bostock2011d3}. 
\subsection{Visual analytics: case study on {\TaskR}}

The model builders and model users can use the {\GNNViz} visualization system to interpret why a graph-based model produces certain correct or incorrect predictions. This is considered as the factual explanation use case. In addition, there is also a counterfactual explanation use case. That some other users may be more interested in the adversarial robustness of a graph-based model, thus they want to identify the vulnarability points of a GNN model through attacking its graph topology. In this section, we use one example (Fig.\ref{fig:subgraphs-short}) to illustrate how {\GNNViz} system can support the multiple user scenarios. In Appendix, we attach more example images. And upon paper acceptance, {\GNNViz} will be publicly available. 

As shown in Fig.\ref{fig:subgraphs-short}, we present one image's three modes (each in a column). The first column presents the raw image, its groundtruth label, and its prediction result generated by the GCN model~\cite{Chen2019MultiLabelIR} that {\GNNViz} aims to explain. Sometimes GCN model may generate incorrect predictions. In this example, the GCN model fails to predict the \texttt{Handbag} label, but mis-predicts an extra \textit{Car} label. 


\paragraph{How does the graph topology contribute to a graph-based model's correct or incorrect prediction results?}

In the \texttt{Preserve} mode, our interpretation algorithm preserves a sparse subgraph (by default ten edges) to reduce the complexity while still maintaining a relative good performance. By looking only at the raw images column and the \texttt{preserve} mode column, we can see that for the data instance, the \texttt{preserve} mode subgraph successfully captures the most important edge \texttt{Person}--to--\texttt{Motorcycle} in the topology (thickness). But the GCN model fails to predict \texttt{Handbag}, as the edge is not in the top-ten edges in the preserved subgraph.
It also mis-predicts the \texttt{Car} label. And the preserve subgraph reflects such misprediction as it believes the \texttt{Person}--to--\texttt{Car} edge has a high importance level. The user can further dive into the graph topology and the CNN classifier to examine why this mistake happens. 

\paragraph{How to improve a graph-based model's prediction performance?}
In the \texttt{Promote} mode, our interpretation algorithm suggests new edges to add to the graph to improve the prediction correctness. 
The GCN model fails to predict \texttt{Handbag}. Our explanation algorithm correctly suggests to add a \texttt{Chair}--to--\texttt{Handbag} link to put more weights on \texttt{Handbag} related topology. And if the user agrees with such suggestion and clicks on that edge in the subgraph, the GCN model will recalculate the prediction result and successfully predict the \texttt{Handbag} label.
{\GNNViz} set parameter to suggest only one edge to add, but the user can use UI to interactively change such setting. 


\paragraph{How to attack a graph-based model?}
In the \texttt{Attack} mode, our interpretation algorithm can suggest edges to be added or deleted from the global graph. The attack mode also targets the global graph topology. 
We use the green dash line to indicate the edges suggested to add, and the red solid line to indicates the edges to be removed. 
By default, {\GNNViz} suggests only one edge to add and one edge to remove, but the user can change this hyperparameter in the control panel via the UI.
As shown in Fig.\ref{fig:subgraphs-short}, the GCN model initially correctly predicts \texttt{Motrocycle} and \texttt{Person}. The {\GNNViz} algorithm suggests to add the \texttt{WineGlass}--to--\texttt{Motorcle} link and remove the \texttt{Handbag}--to--\texttt{Person} link from the graph topology. If the user applies such suggestions, the GCN model will recalculate the prediction result and mistakenly believes \texttt{Wine Glass} is the top prediction label. 

\section{Conclusion}
In this paper, we develop a multi-purpose system for interpreting topology attribution of GNN. The underlying algorithm provides a unified formulation for factual explanation, counterfactual explanation, and universal explanation.
The effectiveness of our proposal has been demonstrated in   GNN-based image classification and social bias detection. 

\clearpage
\newpage

{\small
\bibliographystyle{ICCV2021/ieee_fullname}
\bibliography{refs,ref_SL}
}
\clearpage
\newpage

\setcounter{section}{0}
\setcounter{figure}{0}
\makeatletter 
\renewcommand{\thefigure}{A\@arabic\c@figure}
\makeatother
\setcounter{table}{0}
\renewcommand{\thetable}{A\arabic{table}}
\setcounter{mylemma}{0}
\renewcommand{\themylemma}{A\arabic{mylemma}}
\setcounter{equation}{0}
\renewcommand{\theequation}{A\arabic{equation}}

\appendix

\section{Additional experiment details}
\begin{figure}[h!]
    \centering
    \includegraphics[width=0.95\columnwidth]{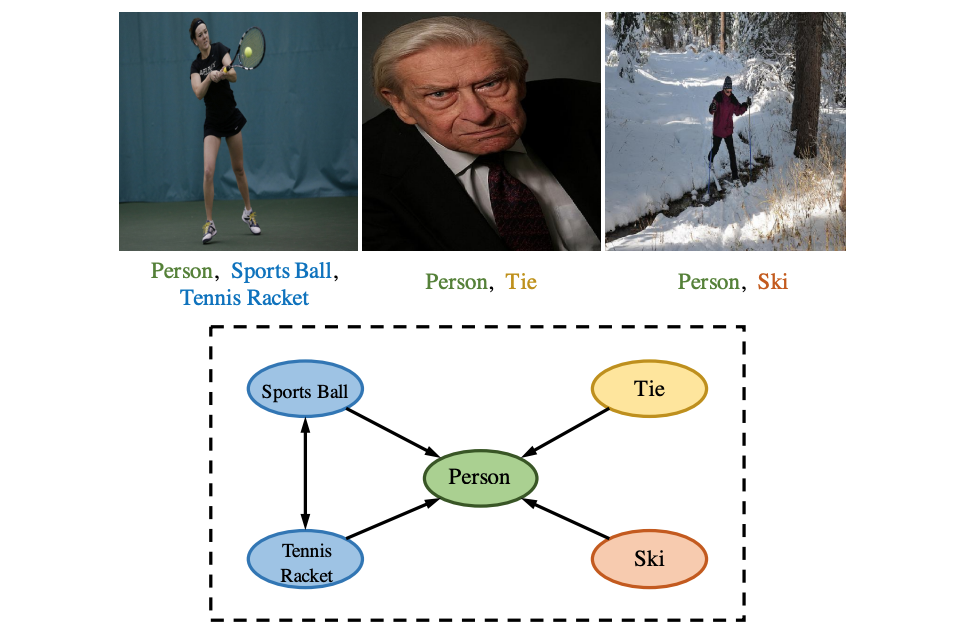}
    \caption{
    The label dependency map illustration for MS-COCO dataset from the original GCN algorithm paper~\cite{Chen2019MultiLabelIR}. Label\_A to Label\_B means when Label\_A appears, Label\_B is likely to appear. The original paper used 0.4 as the threshold for the conditional probability to binarize this topology graph, and we followed their parameter settings.}
    \label{fig:gcn-graph}
\end{figure}

For MS-COCO dataset, we use a well-cited GCN-based multi-label classification algorithm~\cite{Chen2019MultiLabelIR} as the target model for explanation, and demonstrate our interpretation algorithm and the {\GNNViz} system, despite the interpretation algorithm and {\GNNViz} system are applicable to other GNN algorithms and datasets. The global graph topology map in this case is calculated from the label co-occurence probability from the entire training dataset, as shown in Fig.~\ref{fig:gcn-graph}. Thus, each node in the graph represents a label, and each directed edge represents an co-occurence relationship. The edges are binary, with a threshold of conditional probability at 0.4, as suggested by the original paper. A GCN is used to map this global graph to a set of inter-dependent object classifiers. 

An useful explanation in this application is to identify the edges in the global graph that connect labels of interests so the users could understand when certain labels are predicted together. 
We define a reasonable explanation mask as the one that connects relevant labels within $k$-hops. 
Formally, the $k$-hop neighbors of adjacency matrix $\mathbf A$ is defined as $(\mathbf A)^k$.
Given the prediction label set $\Omega$,
we can define the set of relevant edges as 
$
\mathcal S_{\mathrm{rel}} = \{(\mathbf A)^k_{i,j}=1, i\in \Omega, j\in \Omega\}
$
and the set of edges in perturbed graph as
$
\mathcal S_{\mathrm{hit}} = \{(\mathbf A'(\mathbf S)^k_{i,j}=1, i\in \Omega, j\in \Omega\}
$
\paragraph{Evaluating \texttt{preserve} mode}
We measure the explanation quality as the percentage of relevant edges detected by the algorithm, 
$
\frac{|\mathcal S_{\mathrm{hit}}|}{|\mathcal S_{\mathrm{rel}}|}
$
As mentioned in the Section 5.2, the sparse explanation graph in \texttt{preserve} mode is able to identify 62.8\% of the relevant edges connecting  predicted labels.  
\paragraph{Evaluating \texttt{promote} mode}
One of the assumption is that suggested edge additions increase probability of true labels by increasing connectivity between true labels in the adjacency matrix. To verify it, we calculate the $P$-value of the influence of recommended perturbed graph on GNN's prediction against randomly perturbed graphs.
Let $m$ be the number of walks between true labels after adding top $k$ of the recommended edges. Let $n_i$ be the   number of walks between true labels after randomly adding $k$ random edges outside the adjacency matrix (we call this a realization of a random mask). The $P$-value is then given by the probability of significance on how the real observation $m$ is located among the random observations $\{ n_i\}_{i=1}^{1000}$ for $1000$ random trials, namely, $P(m) = \mathrm{Prob}(m \geq R_n)$, where $R_n $ is a random variable uniformly drawn from $\{ n_i\}$.

\section{{\GNNViz} system overview}

\begin{figure*}[htb]
\vspace*{-0.0in}
\centerline{
\begin{tabular}{c}
\includegraphics[width=0.99\textwidth,height=!]{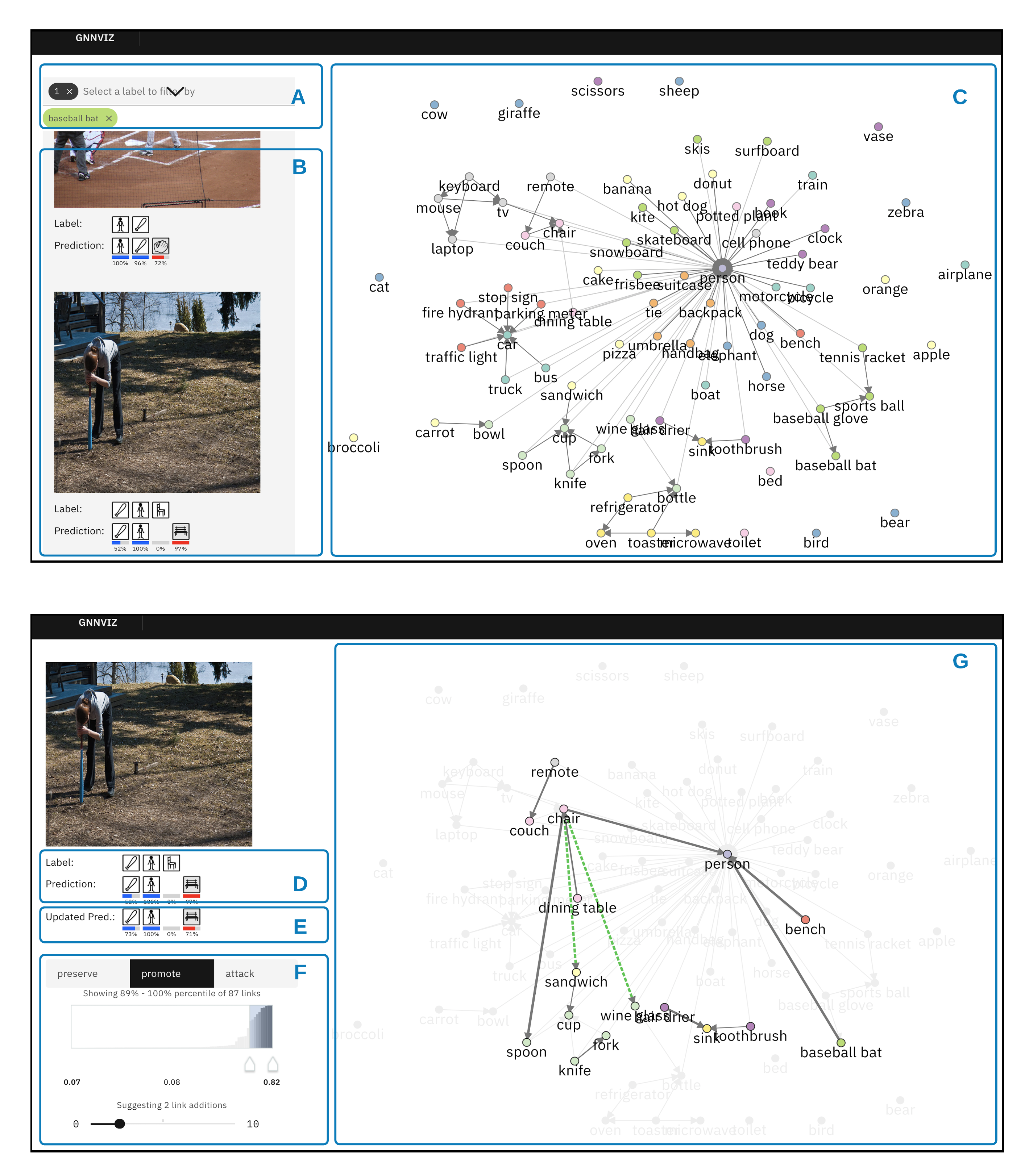}
\end{tabular}}
\caption{{The User Interface of {\GNNViz}. It has two views: a browser view (TOP), and a detailed view for a single data instance (BOTTOM). 
}}
\label{fig:system-ui}
\vspace*{-0.00in}
\end{figure*}

We design {\GNNViz} as an interactive and multi-purpose visualization system for interpreting GNN models, as shown in Fig.\ref{fig:system-ui}. In this section, we elaborate its two user views: a browser view (Fig.\ref{fig:system-ui} \textbf{Top}), and a detailed view (Fig.\ref{fig:system-ui} \textbf{Bottom}).

\paragraph{Browser View} The browser view is the landing page for a user. It consists with three main sections: a filter (Fig.\ref{fig:system-ui} \textbf{A}) that allows a user to specify the filtering condition based on the images' ground truth labels; a navigation list (Fig.\ref{fig:system-ui} \textbf{B}) that allows users to browse all the data instances; and a graph canvas (Fig.\ref{fig:system-ui} \textbf{C}) that presents the global graph learned from the entire dataset, which is agnostic to individual data instances.

The main function of the browser view is to allow a user to quickly navigate through a large amount of data instances and identify the data instances that may be of interest to the user. As shown in Fig.\ref{fig:system-ui} \textbf{B}, the list contains each image, and its groundtruth labels and the targeted GCN algorithm~\cite{Chen2019MultiLabelIR} predicted labels. Noted that the targeted GCN algorithm can have incorrect predictions. For example, in the presented image example in (Fig.\ref{fig:system-ui} \textbf{B}), GCN failed to predict the chair label, but add an extra bench label, when compared to the groundtruth labels. 

Different user groups may have different user needs for interpretation, thus the system allows each individual user to choose which data instances may be more of interest to them. For example, an non-expert model user may be interested in understanding only the correct prediction cases, thus he or she can click on the images that have an accurate prediction result; whereas an expert GNN model builder may want to understand why the examined GCN algorithm always fails to predict certain labels, thus he or she can click on those images to explore further details. 

The filter function in Fig.\ref{fig:system-ui} \textbf{A} is to further ease such user navigation. In the demonstrated MS-COCO dataset, there are hundreds of thousands image data instances, and it is really hard for a user to manually browse through all the images, thus they can use the filter function to limit the UI to show only a subset of images based on their groundtruth labels. The filter function also support AND operation when a user enters more than one label as the filtering criteria, thus it can further helps users to navigate through large datasets like the demonstrated MS-COCO.


The majority space of the browser view UI is dedicated to the global graph topology. This global graph topology can be considered as equivalent to a knowledge graph that contains the domain knowledge of the MS-COCO dataset. By nature, this knowledge graph is agnostic to individual data instances. In the original paper~\cite{Chen2019MultiLabelIR}, the authors only presented an illustrative graph with 5 nodes (Fig. \ref{fig:gcn-graph}). {\GNNViz} presents the entire graph topology with all 80 labels as nodes. The color of nodes are defined by the node categories in the MS-COCO dataset. For example, bird, giraffe, cow and a few other nodes are colored in blue as they all belong to the animal category; and the sport category nodes (baseball bat, skateboard, frisbee, etc) are colored in light green.

The layout of this graph (Fig.~\ref{fig:system-ui} \textbf{C}) uses a force-directed graph layout~\cite{fruchterman1991graph} and the implementation uses a D3.js javascript library~\cite{bostock2011d3}. The rendering algorithm first identifies the central node by calculating the graph's centrality and the degree of the nodes (person node in this case), then distribute other nodes based on their connectivity to the central nodes and their degrees. The disconnected nodes are automatically pushed to the peripheral areas. This global graph is the foundation of the targeted GCN algorithm for interpretation, and it can already convey some meaningful information, such as the oven, refrigerator, toaster and other kitchen appliance are often appear together in an image, whereas giraffe or bear is often alone in an image in the MS-COCO dataset.

\paragraph{Detailed View} After navigating through the browser view (Fig.\ref{fig:system-ui}Top) and identifying a particular image of interest, a user can go to an image's detailed view (Fig.\ref{fig:system-ui}Bottom) by clicking on that image. The detailed view has four major areas, in addition to the selected raw image: the groundtruth label and the GCN prediction result (Fig.\ref{fig:system-ui}\textbf{D}), the updated prediction results (Fig.\ref{fig:system-ui}\textbf{E}) after the user apply one of the three user actions (i.e., \texttt{preserve}, \texttt{promote}, and \texttt{attack} modes), a user control panel that a user can switch between the three user actions and manipulate hyperparameters (Fig.\ref{fig:system-ui}\textbf{F}), and the graph topology canvas (Fig.\ref{fig:system-ui}\textbf{G}). 

The default user mode is \texttt{preserve} mode, which is not shown in Fig.\ref{fig:system-ui}. The \texttt{preserve} mode falls into the factual explanation use case -- the explanation algorithm identifies a less complex subgraph for prediction, but still try to maintain the best prediction performance as using the original global graph topology. In the user control panel (Fig.\ref{fig:system-ui}\textbf{F}), users can adjust the percentile range selector to revise how many edges they want to keep in the subgraph, which will further impact the \texttt{updated prediction} results in Fig.\ref{fig:system-ui}\textbf{E}. The less edges a user keeps in the subgraph, the worse updated prediction result is. By default, the algorithm identify the subgraph with 10 edges. All rest of the unpreserved edges are shown in a 10\% opacity in the background. In the following visual analytics subsection, we will present more details of this \texttt{preserve} explanation mode.

The second user mode is \texttt{promote} mode, and it is illustrated in Fig.\ref{fig:system-ui}. The \texttt{promote} mode is also for the factual explanation user scenario. For the targeted GNN prediction algorithm (GCN from ~\cite{Chen2019MultiLabelIR} in this case), our {\GNNViz} algorithm can suggest new edges to add into the graph topology and promote the correct prediction results. In the control panel in Fig.\ref{fig:system-ui}\textbf{F}, a user can adjust the progress bar at the bottom to decide how many additional edges they want the {\GNNViz} algorithm to suggest. For example, in Fig.\ref{fig:system-ui}\textbf{Bottom}, the user sets the parameter to two edges for addition. In Fig.~\ref{fig:system-ui}\textbf{G}, the suggested two edges appears in green dash lines (chair to sandwich and chair to wine glass). If a user agrees with the recommendation and wants to explore this suggestion, they can simply click on the green dash lines in Fig.~\ref{fig:system-ui}\textbf{G} to add those edges into the graph topology. The GCN prediction algorithm will recalculate the prediction result using this manipulated graph topology, and the prediction results will be updated in Fig.~\ref{fig:system-ui}\textbf{E}. By default, \texttt{promote} suggests one edge to add. Later in the visual analytics subsection, we will showcase the effectiveness of the proposed {\GNNViz} algorithm with a few data instances.

\begin{figure*}[h]
\vspace*{-0.0in}
\centerline{
\begin{tabular}{c}
\includegraphics[width=0.97\textwidth,height=!]{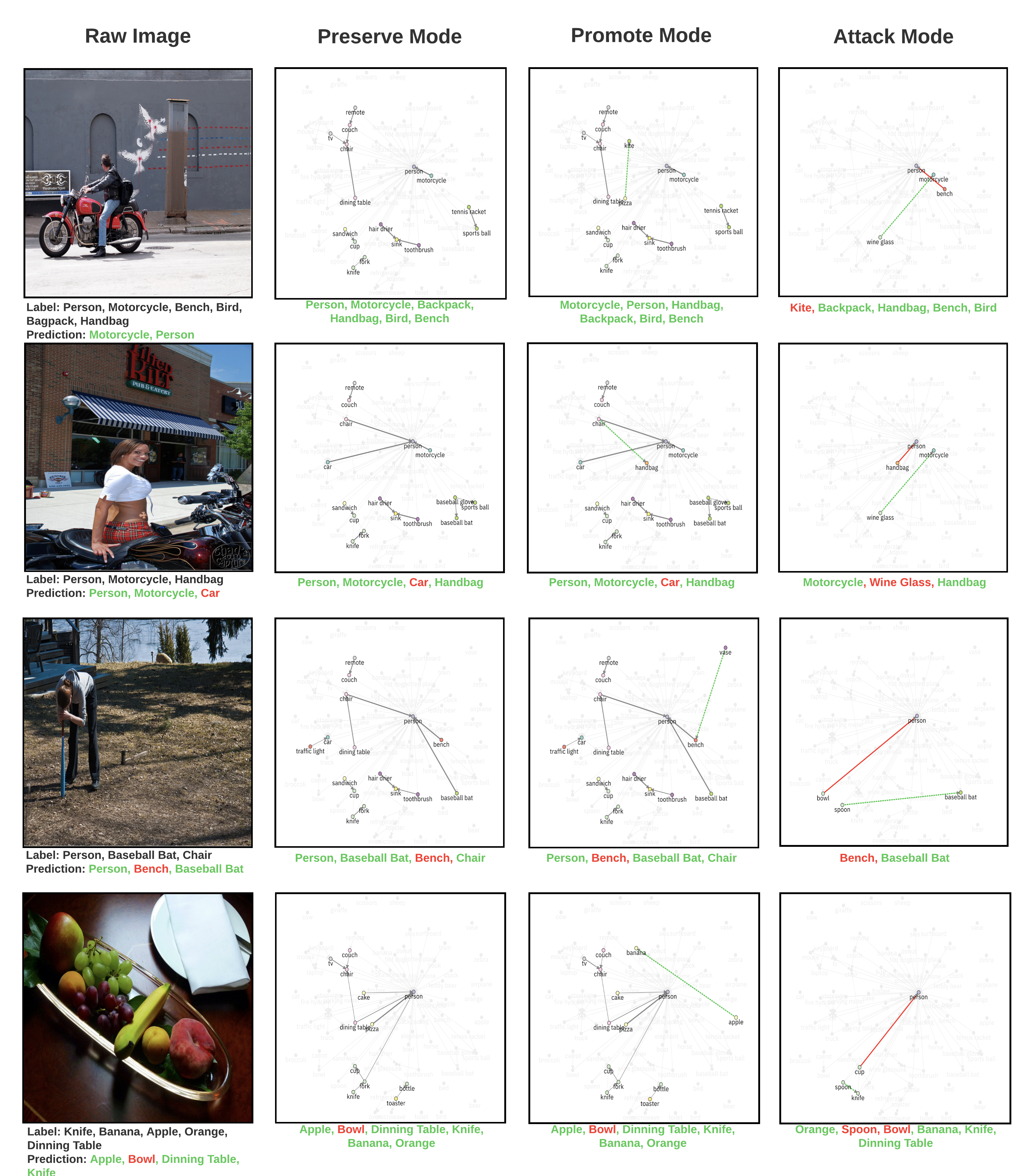}
\end{tabular}}
\caption{{Each row represents one data instance and its \texttt{preserve} mode, \texttt{promote} mode, and \texttt{attack} mode.}}
\label{fig:subgraphs}
\vspace*{-0.2in}
\end{figure*}

The third user mode is \texttt{attack} mode. In contrary to the previous two user modes, this mode is to visualize the counterfactual explanation user scenarios. {\GNNViz} can recommend the minimal edge changes (both edge addition and deletion) to attack the GCN prediction algorithm -- manipulate the updated prediction results to move away from the correct predictions. The user interface looks similar to the \texttt{promote} mode shown in Fig.~\ref{fig:system-ui} \textbf{Bottom}. The difference between \texttt{promote} mode and \texttt{attack} mode is that in the graph canvas Fig.~\ref{fig:system-ui} \textbf{G}, the \texttt{attack} mode hide the entire graph topology in the background and highlight only the edges recommended for attacking. It uses green dash lines to indicate the edge addition, and uses red solid lines to indicate edge removal, as shown in Fig.~\ref{fig:subgraphs}.

In summary, the proposed {\GNNViz} system provides an easy-to-use and interactive visualization system for both graph-based model experts and non-technical model users to visually interpret, analyze, and attack GNN models. The implemented three modes --- \texttt{preserve, promote, and attack} --- satisfy both the factual explanation use case and the counterfactual explanation use case. The visualization system architecture is implemented as a general purpose API-based web service, thus researchers can easily switch the interpretation backend to their own interpretation algorithms and visually interpret its usefulness. Also, if the researcher is not interested in GNN interpretation research topic, the visualization system can simply be used to examine a graph-based model's prediction performance at the instance level.

\section{Qualitative visual analytics with MS-COCO as a case study}

As noted in the previous subsection, the model builders and model users can use the {\GNNViz} visualization system to interpret why a graph-based model produces certain predictions, both correct and incorrect ones. This is considered as the factual explanation use case. In addition, there is also a counterfactual explanation use case. That some other users may be more interested in the adversarial robustness of a graph-based model, thus they want to identify the vulnarability points of a GNN model through attacking its graph topology. In this section, we will use a few examples (Fig.\ref{fig:subgraphs}) to illustrate how our implemented {\GNNViz} algorithm and visualization system support both use cases.

As shown in Fig.\ref{fig:subgraphs}, we present four image data instances (each in a row) and their three modes (each in a column). In the first column, we present the raw image, its groundtruth label, and its prediction result generated by the GCN model~\cite{Chen2019MultiLabelIR} that {\GNNViz} aims to explain. Sometimes the GCN prediction model may generate incorrect predictions. For example, in the second row of the data, the GCN model fails to predict the \texttt{Handbag} label, but mis-predicts an extra \textit{Car} label. For the labels that GCN correctly predicts, a user may be interested to know why the graph-based model makes such correct prediction; for the incorrect labels, a user may want to know why the model makes a wrong prediction. These user needs are categorized as \textbf{factual explanation} use cases, and we design a \texttt{Preserve} mode and a \texttt{Promote} mode for it. In addition, a user may wants to increase the robustness of a graph-based model, thus he or she needs to identify the adversarial vulnerability of the graph topology--where in the graph a user can spent the minimal effort to change the edges and to negatively impact the prediction result. We treat this use case separately as a \textbf{counterfactual explanation} need and design the \texttt{Attack} mode for it. 

In summary, we will use the rest of this subsection to illustrate how {\GNNViz} visualization system can help a user to understand, improve, and attack a model's prediction outcome by examining and interacting with the visual graph topology. 

\paragraph{\texttt{Preserve} explanation mode}

In the \texttt{preserve} mode, our interpretation algorithm preserves a sparse subgraph (by default ten-edges) to reduce the complexity while still maintaining a relative good performance. By looking only at the raw images column and the \texttt{preserve} mode column, we can see that for the data instance (1), the \texttt{preserve} mode subgraph successfully captures the most important edge \texttt{Person}--to--\texttt{Motorcycle} in the topology, because the \texttt{Person}--to--\texttt{Motorcycle} link is the thickest one. But the GCN model fails to predict \texttt{Bird}, \texttt{Bagpack} or \texttt{Handbag}, as these edges are not in the top-ten edges in the preserved subgraph.

In contrary, we can see from the second data instance (2) that the GCN model mis-predicts the \texttt{Car} label. And the preserve subgraph reflects such misprediction as it believes the \texttt{Person}--to--\texttt{Car} edge has a high importance level. The model builder can further dive into the graph topology and the CNN classifier to examine why this mistake happens. 

\paragraph{\texttt{Promote} explanation mode}
In the \textit{promote} mode, our interpretation algorithm suggests new edges to add to the graph to improve the prediction correctness. Worth to mention, the promote mode is targeting the global graph topology, instead of the preserved subgraph. The user can click on the suggested edges in the subgraph in the \textit{promote} mode and the selected edges will be added onto the graph topology, and the prediction probabilities will be recalculated and updated. By default and as illustrated in Fig.\ref{fig:subgraphs}, we set the parameter to allow {\GNNViz} algorithm to suggest only one edge to add, but the user can use the control panel in Fig.\ref{fig:system-ui}\textbf{F} to interactively change such setting. 

For example, for the fourth data instance, the GCN model fails to predict \texttt{Banana} and \texttt{Orange}. As shown in the \texttt{Promote} mode, the {\GNNViz} algorithm correctly suggest to add a \texttt{Banana}--to--\texttt{Apple} link. And if the user agrees with such suggestion and clicks on that edge in the subgraph, the GCN model will recalculate the prediction result and successfully predict the \texttt{Banana} and \texttt{Orange} labels, while not affecting other labels.

\paragraph{\texttt{Attack} explanation mode}
In the \texttt{attack} mode, the interpretation algorithm can suggest both edges to be added and edges to be deleted. The attack mode also targets the global graph topology, thus the edges to be added or to be deleted are selected from the global graph. We use the green dash line to indicate the edges suggested to add, and the red solid line to indicates the edges to be removed. By default, we set the parameter for {\GNNViz} algorithm to suggest only one edge to add and one edge to remove, but the user can change this hyperparameter in the control panel via the UI.

As shown in the first data instance, the GCN model initially correctly predicts \texttt{Motrocycle} and \texttt{Person}. The {\GNNViz} algorithm suggests to add the \texttt{WineGlass}--to--\texttt{Motorcle} link and remove the \texttt{Bench}--to--\texttt{Person} link from the graph topology. If the user applies such suggestions, the GCN model will recalculate the prediction result and mistakenly believes \texttt{Kite} is the top prediction label. Similarly, in the third data instance row, if the user agrees with the algorithm suggestion by adding \texttt{Spoon}--to--\texttt{BaseballBat} and removing \texttt{Bowl}--to--\texttt{Person} edges, the GCN model can not predict the \texttt{Person} label anymore, and the attack is successful.

\end{document}